# ADAM: An AI Reasoning and Bioinformatics Model for Alzheimer's Disease Detection and Microbiome-Clinical Data Integration

**Ziyuan Huang[1,4,5], Vishaldeep Kaur Sekhon[2], Roozbeh Sadeghian[3], Maria L. Vaida[3], Cynthia Jo[4], Beth A. McCormick[1,5], Doyle V. Ward[1,5], Vanni Bucci[1,5], & John P. Haran[1,4,5]**

[1] Department of Microbiology, UMass Chan Medical School, Worcester, MA, USA
[2] Department of Geriatric Medicine and Gerontology, Johns Hopkins University, Baltimore, MD, USA
[3] Data Sciences, Harrisburg University of Science and Technology, Harrisburg, PA, USA
[4] Department of Emergency Medicine, UMass Chan Medical School, Worcester, MA, USA
[5] Program in Microbiome Dynamics, UMass Chan Medical School, Worcester, MA, USA

Corresponding author: Ziyuan Huang (e-mail: ziyuan.huang2@umassmed.edu).

This work was supported by the National Institutes of Health (NIH) under Grant R01AG067483-01. The project was conducted within the Haran Research Group at the University of Massachusetts Chan Medical School, under the direction of Principal Investigator John P. Haran.

**ABSTRACT** Alzheimer's Disease Analysis Model (ADAM) is a multi-agent reasoning large language model (LLM) framework designed to integrate and analyze multimodal data, including microbiome profiles, clinical datasets, and external knowledge bases, to enhance the understanding and classification of Alzheimer's disease (AD). By leveraging the agentic system with LLM, ADAM produces insights from diverse data sources and contextualizes the findings with literature-driven evidence. A comparative evaluation with XGBoost revealed a significantly improved mean F1 score and significantly reduced variance for ADAM, highlighting its robustness and consistency, particularly when utilizing human biological data. Although currently tailored for binary classification tasks with two data modalities, future iterations will aim to incorporate additional data types, such as neuroimaging and peripheral biomarkers, and expand them to predict disease progression, thereby broadening ADAM's scalability and applicability in AD research and diagnostic applications.

## I. INTRODUCTION

The integration of multimodal data sources in biomedical research has accelerated with advances in deep learning, particularly through the development of large language models (LLMs). The introduction of AlexNet in 2012 demonstrated the potential of deep neural networks for complex tasks such as image classification [1]. Subsequent developments, including transformer architecture in 2017 and the release of models like GPT-1 in 2018 and GPT-2 in 2019, marked significant milestones in natural language processing [2, 3]. Other transformer-based LLM families, such as LLaMA, Gemini, Claude, and DeepSeek, have also demonstrated powerful structured and unstructured biomedical data processing [4-10]. More recently, reasoning-focused LLMs, including GPT-4.5, OpenAI o1, Gemini 2.5, Claude Sonnet 3.7, and DeepSeek R1, have further extended the role of artificial intelligence (AI) in scientific domains by enabling more complex inference and task coordination [11-15].

In bioscience, LLMs have shown promise in clinical decision support, patient-trial matching, and biomedical question answering [16-18]. Concurrent developments like AlphaFold and ESM-2 have led to protein structure prediction and variant interpretation breakthroughs [19, 20]. Frameworks like BioLunar, ASD-cancer, and CHIEF exemplify the value of integrating multimodal evidence, including genomic, imaging, and text-based modalities, for enhanced diagnostics in cancer and other diseases [21-23].

While cancer research has rapidly integrated AI tools across diagnostics, treatment planning, and drug discovery, research into Alzheimer's disease (AD) has been slower to adopt these technologies on a comparable scale. AD is a multifactorial neurodegenerative disorder characterized by beta-amyloid plaques, tau protein tangles, immune system dysregulation, and changes in the gut microbiome [24, 25]. Traditional studies often depend on single-modality datasets, which can limit comprehensive insights into the disease's progression.

This study presents the Alzheimer’s Disease Analysis Model (ADAM), a multi-agent reasoning framework leveraging large language models (LLMs) and retrieval-augmented generation (RAG) [26]. ADAM combines microbiome and clinical data to improve the analysis of Alzheimer's disease. The current version, called Generation 1 (ADAM-1), integrates modular agents with Chain-of-Thought (CoT) [27] reasoning to enhance interpretability and contextual relevance in disease classification. Trained and tested on a laboratory dataset of 335 multimodal data samples from older adults, the framework achieved a significantly higher mean F1 score and a much lower prediction variance than the XGBoost baseline model. These findings highlight the methodological benefits of agentic reasoning systems in maintaining performance under data-limited situations.

## II. Multimodal Dataset Description and Visualization

This study utilizes the nursing home clinical and metagenomic dataset originally compiled by Haran et al. [28], which focused on the dysregulation of the anti-inflammatory P-glycoprotein pathway in AD. The dataset is repurposed to develop ADAM, a large LLM-based classifier and reporting system in this work. Its representative size is typical of datasets routinely produced. The dataset contains clinical and microbiome data from nursing home residents in central Massachusetts, focusing on individuals with and without AD. The analysis included 335 stool samples collected from approximately 100 unique participants, with some individuals providing multiple samples over up to five months.

### *A. Clinical Data*

Within the clinical dataset, each sample was annotated with detailed clinical metadata, including AD status, comorbidities, demographic information, and longitudinal sampling data. Of the total sample, 32.84% came from individuals diagnosed with AD, while the remaining 67.16% came from individuals without dementia. The cohort is predominantly female (85.7%), with a mean age of 84.5 years and a median of 86.0 years, spanning an age range from 52 to 102 years. This age distribution reflects the demographic characteristics typical of long-term care populations (FIGURE 1).

In addition to cross-sectional clinical features, the dataset included a longitudinal sampling component. Participants contributed between one and 12 stool samples, with a median of three samples per participant, enabling within-subject comparisons over time. Sampling timelines varied by individual, with consistent representation across the AD and non-AD groups. This temporal resolution supports dynamic analyses of microbiome profiles and disease status. The integrated design of the dataset, linking clinical diagnoses, demographic characteristics, comorbid conditions, and longitudinal biospecimens, provides a robust framework for exploring the complex interactions between neurodegenerative diseases, aging, and host-associated microbial communities. These clinical features were integrated into the reasoning and analytical processes of the ADAM framework.

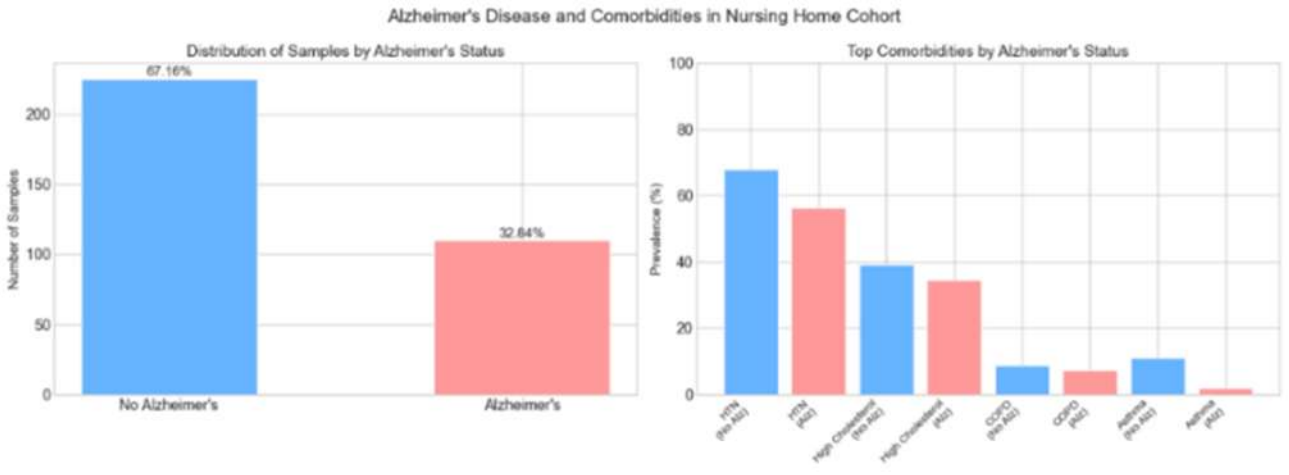

**(A) Alzheimer’s Disease and Comorbidities in Nursing Home Cohort**

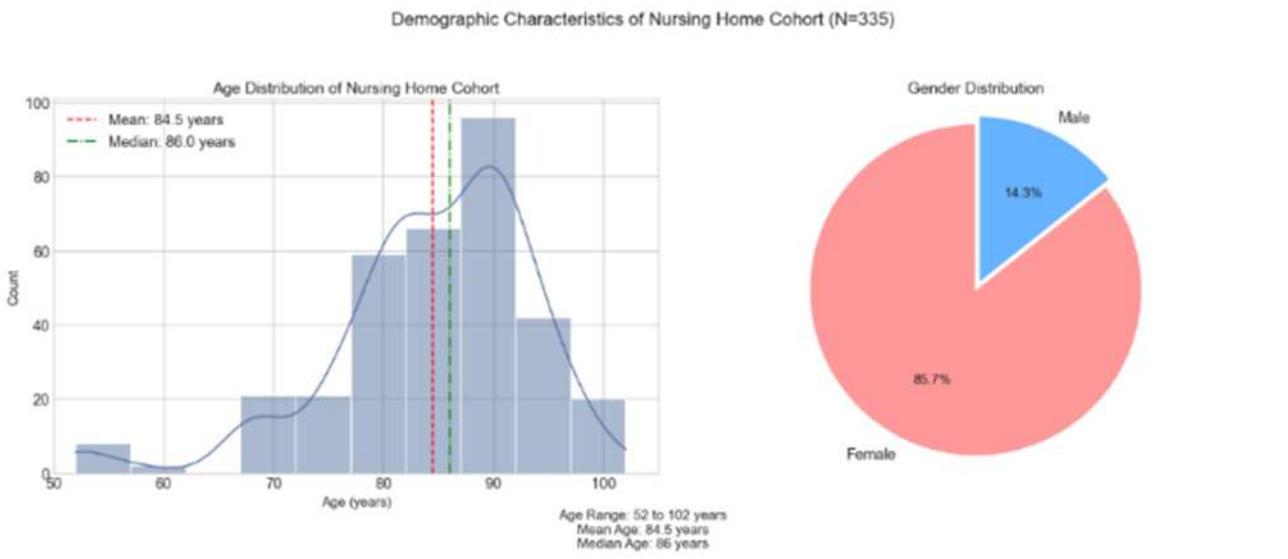

**(B) Demographic Characteristics of Nursing Home Cohort**

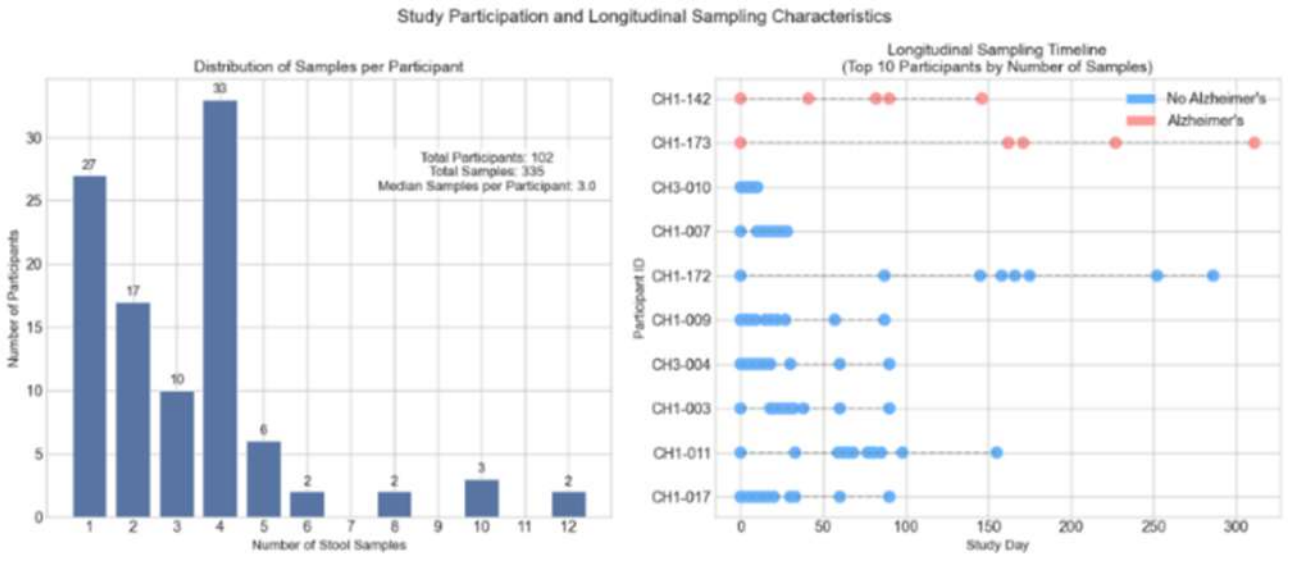

**(C) Study Participation and Longitudinal Sampling Characteristics**

**FIGURE 1. The Nursing Home Cohort Clinical Data Description (source: [28]). (A) Bar plots illustrate the distribution of participants based on Alzheimer's status and the prevalence of key comorbidities. 67.16% of participants do not have Alzheimer's disease, while 32.84% do. The most prevalent comorbidities include hypertension (HTN), and various cardiovascular diseases (CVD), with prevalence categorized by Alzheimer's status. (B) The demographic characteristics of the cohort (N=335) encompass age distribution, with a mean age of 84.5 years and a median age of 86.0 years, alongside a pie chart depicting gender distribution, which shows a predominance of females (85.7%) in the cohort. (C) Study participation metrics include the distribution of stool samples per participant and a longitudinal sampling timeline for the top 10 participants based on the number of samples collected. The median number of samples per participant is 3. Alzheimer's status is color-coded across all visualizations**

### B. Gut Microbiome Data

We downsized the gut microbiome dataset to 940 bacterial species based on their prevalence and relative abundance. The original microbiome dataset was generated through a standardized process, beginning with collecting stool samples from nursing home residents, followed by DNA extraction using a Qiagen DNeasy PowerSoil Pro Kit. The resulting DNA was used to create a pool containing 2 nM DNA, 12 µL RSB with Tween, and 4 µL of diluted PhiX prep, which was then pipetted into a P4 Illumina flow cell cartridge. The sequencing run was generated on the BaseSpace Illumina platform and validated using Illumina NextSeq 2000 prior to analysis.

We characterized the dataset by analyzing species-level prevalence, abundance, and differential abundance metrics, as shown in FIGURE 2. The prevalence–abundance distribution (FIGURE 2(A)) exhibits a typical long-tailed pattern, where most bacterial species show low prevalence and low proportional abundance. These rare taxa often add to noise and variance, reducing statistical power. To address this issue, we implemented filtering thresholds, retaining only species present in at least 5% of samples (prevalence ≥17) and with a total abundance greater than 1e-4. This approach ensures that our analysis focuses on reproducible and biologically relevant species.

After preprocessing, we analyzed differential abundance to pinpoint taxa significantly linked to AD status. The volcano plot in FIGURE 2(B) displays $\log_2$ fold changes versus $-\log_{10}$(p-values), indicating species that are higher in AD (red) and those that are lower (blue). This thorough profiling approach improves the signal-to-noise ratio, ensuring that only reliable microbial features contribute to the ADAM framework.

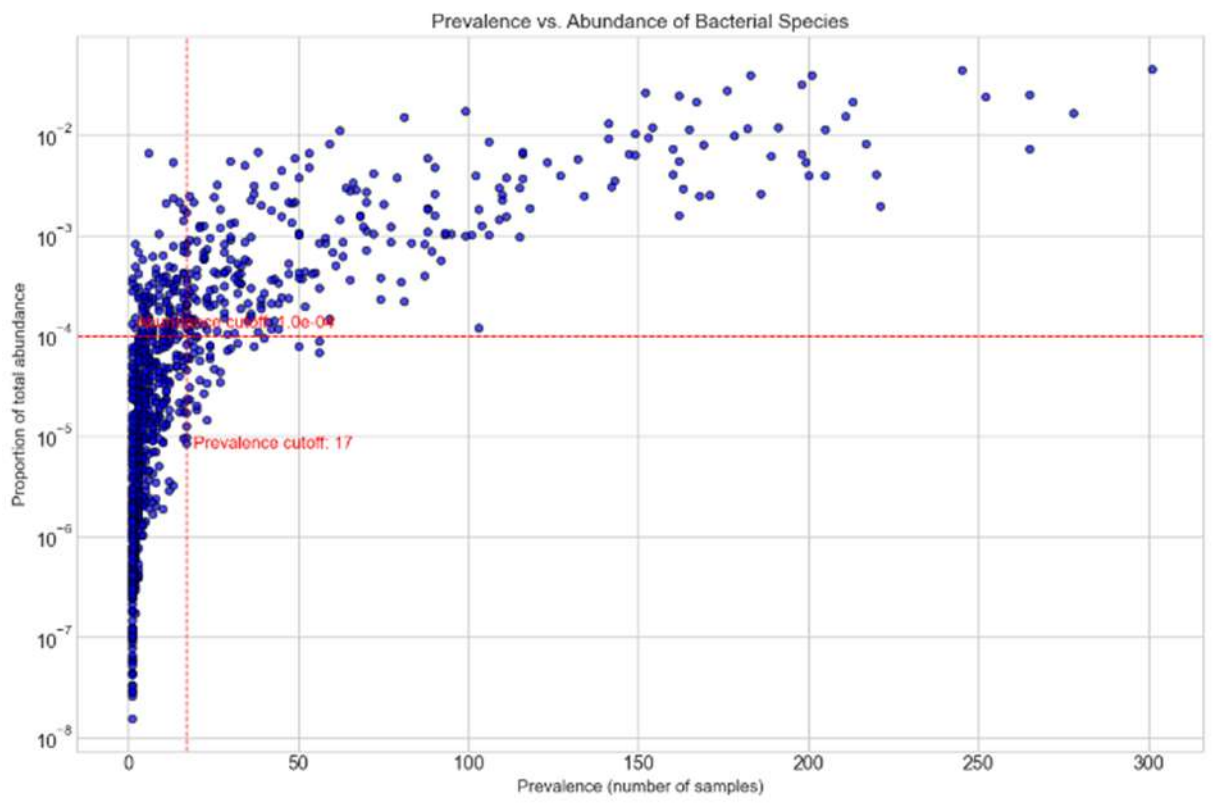

(A) Prevalence vs. Abundance of Bacterial Species

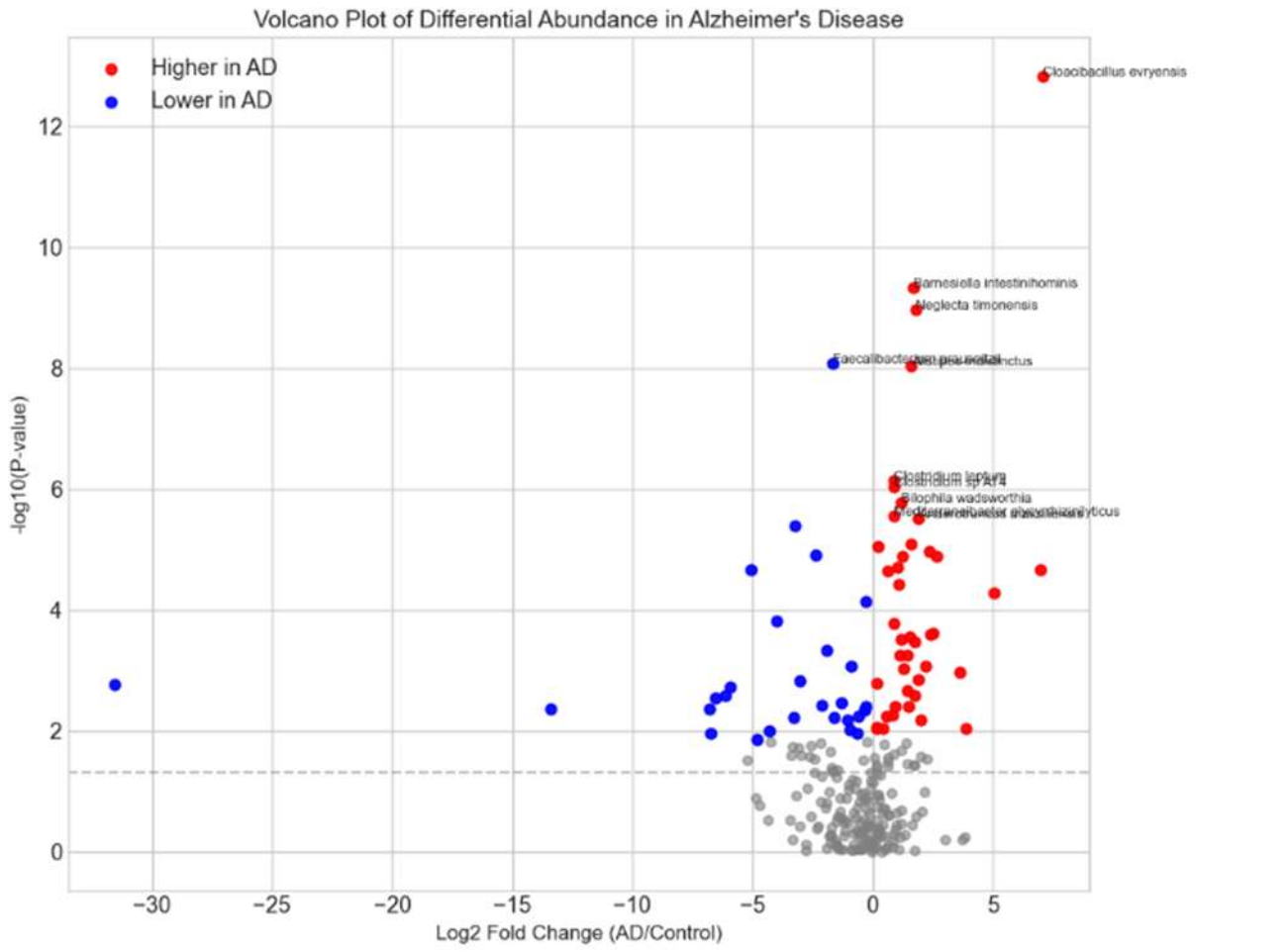

(B) Volcano Plot of Differential Abundance in Alzheimer's Disease

**FIGURE 2. Nursing Home Cohort Bacteria Data Description (source: [28]). (A) Prevalence vs. Abundance of Bacterial Species: Scatter plot illustrating bacterial species by their prevalence (the number of samples in which each species appears) and proportional abundance across the dataset. Red dashed lines represent the filtering criteria applied during preprocessing: species present in fewer than 5% of samples (prevalence cutoff: 17) or with a total relative abundance below 1e-4 were removed to minimize noise and focus on biologically relevant taxa. (B) Volcano Plot of Differential Abundance: $\text{Log}_2$ fold change versus $-\log_{10}$(p-value) for all bacterial species. Species with significantly higher abundance in Alzheimer's patients are shown in red; those lower in AD are shown in blue.**

## III. Architecture of ADAM

The ADAM framework comprises an agentic system, a semantic search engine, two base LLMs, and a reporting module, along with new and existing laboratory data. FIGURE 3 illustrates the architecture of the ADAM framework. The ADAM framework begins with users inputting new data into a system. The agentic system processes the new data and sends the new data and computational results to the summarization and classification agents, along with historical laboratory data. This information is then processed through a semantic research engine and LLM, while the cosine similarity controls the return quality. Finally, a reporting system presents the output as a classification report.

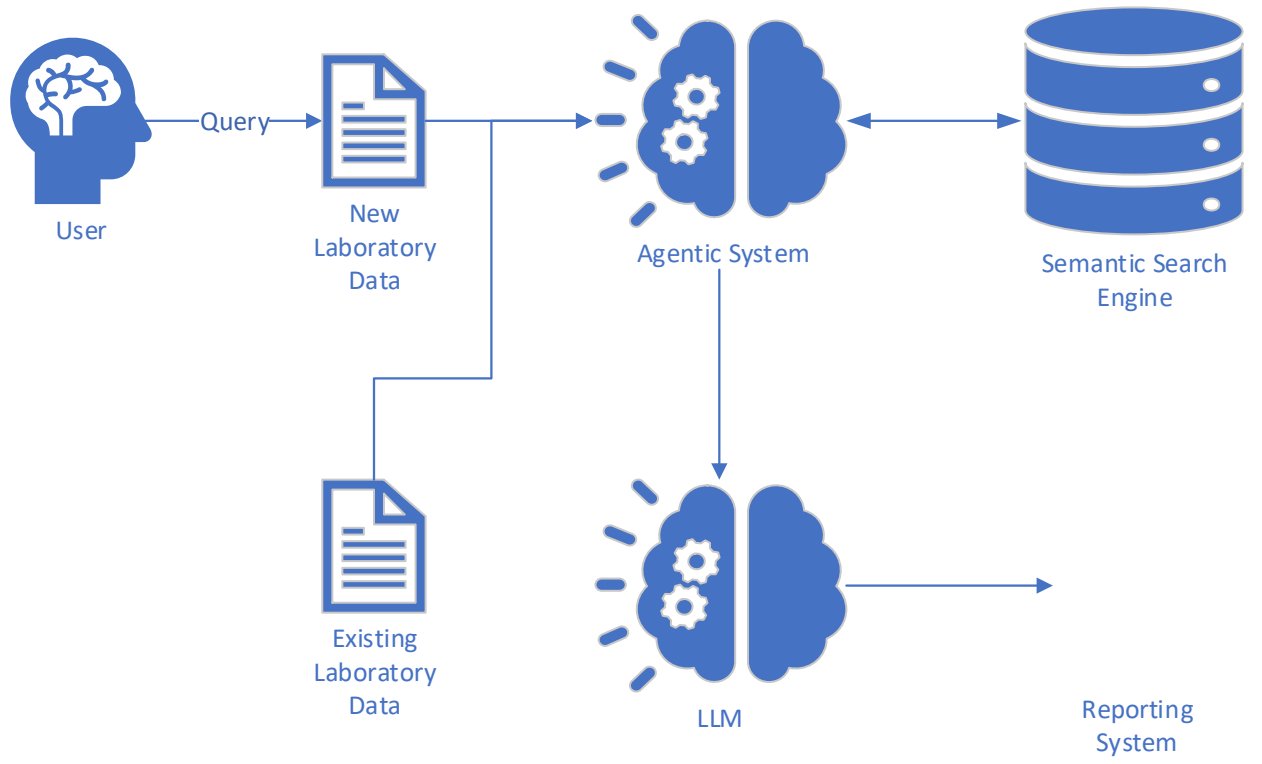

**FIGURE 3. Workflow of the ADAM Framework. This diagram illustrates user interactions within the ADAM system. When a user submits a query along with new laboratory data, the framework employs the agentic system to refine the query and identify relevant literature evidence from a semantic search engine, integrating it with existing laboratory data. The semantic search engine collaborates with the agentic system, alongside**

**the LLM, to reason, analyze, and interpret data, ultimately delivering precise analytical reports.**

### A. Agentic System

The agentic system of the ADAM framework consists of three functionally distinct yet interdependent agents: a computational agent, a summarization agent, and a classification agent. These agents work in a coordinated sequence to transform clinical and metagenomic data into interpretable patient-specific insights. As illustrated in FIGURE 4, the computational agent processes the newly uploaded data to extract key features and analytical outputs. These outputs, along with historical laboratory data, are then utilized by summarization and classification agents, each guided by CoT reasoning and supported by LLMs. The summarization agent generated context-aware narratives, whereas the classification agent determined the AD status.

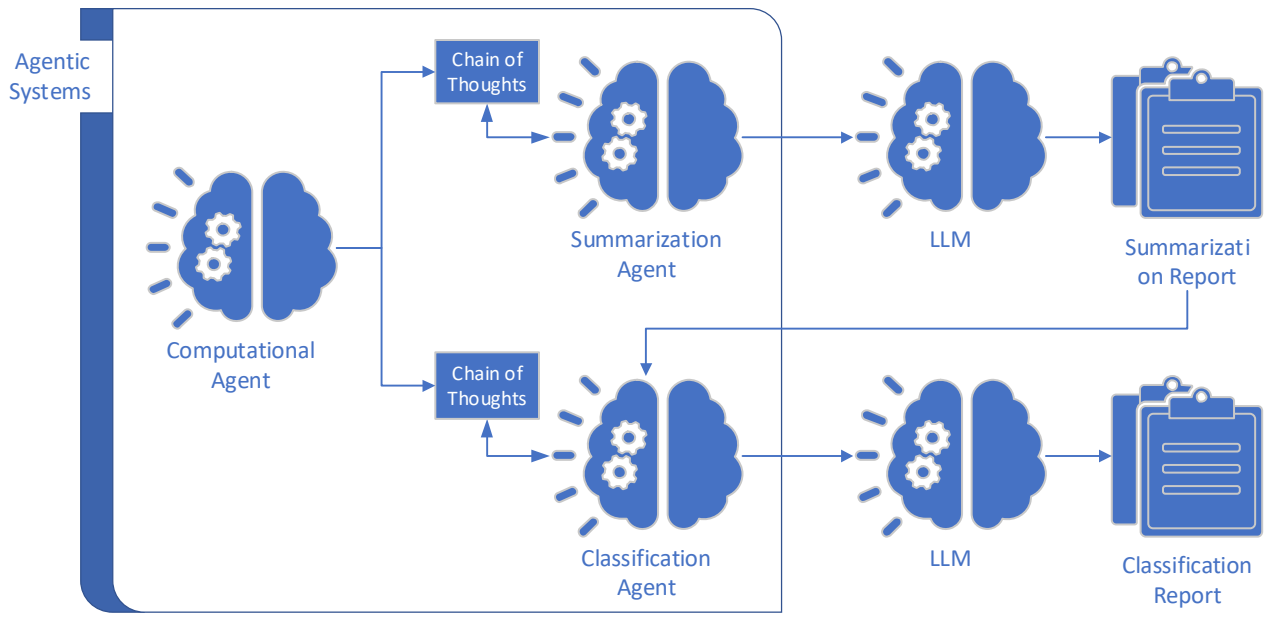

**FIGURE 4. Agentic System Workflow. The Diagram illustrates the coordinated interaction among three AI agents—Computational, Summarization, and Classification—within the agentic system. Each agent leverages Chain-of-Thought reasoning and communicates with the LLM to generate structured outputs. The system collectively produces Alzheimer's-specific summarization and classification reports by integrating computational analysis with contextual interpretation.**

*Computational Agent*: The computational agent comprises bioinformatics and machine learning. Alpha diversity (Shannon, Simpson, Berger-Parker dominance) and beta diversity (Bray-Curtis, Jaccard, Canberra) were calculated to provide a descriptive analysis of bacterial relationships. XGBoost was chosen as the primary machine learning model, working in conjunction with SHAP to generate feature importance information and SHAP values that describe feature interactions in relation to AD status from a global and an individual study subject perspective. The formula for the computational agent is as follows:

$$CA_{comp}(MD, CF) = SHAP\big(XGB(MD, CF)\big), D_{\alpha}(MD), D_{\beta}(MD) \quad (1)$$

Where:

$CA_{comp}$ is the Computational Agent responsible for processing microbiome and clinical data to generate interpretable outputs.

MD is the Microbial Data input (e.g., microbial abundance profiles).

CF is a clinical feature input (e.g., patient demographics and health records).

SHAP(XGB(MD, CF)) represents the XGBoost model embedded within SHAP, which enables interpretable machine learning via feature attributes and interactions.

$D_a$(MD) denotes the Alpha diversity metrics (Shannon, Simpson, Berger-Parker dominance) computed from microbial data.

$D_{\beta}$(MD) denotes the Beta diversity metrics (Bray-Curtis, Jaccard, Canberra) representing microbial community dissimilarity.

*Summarization Agent*: The summarization agent integrates the computational agent's results, CoT reasoning, semantic search engine, and GPT-4o model to provide an overall context summary. The model demonstrates superior language understanding and coherence in long text **[29]**. Its ability to maintain contextual relevance and accuracy in summaries is well documented, particularly in academic and technical domains. This makes it an ideal choice for summarization tasks that require precision and reliability. The formula for the summarization agent is as follows:

$$SA_{summary}\left(CA_{comp}, CoT_{reasoning}, SS_{search}, LLM_{GPT\text{-}4o}\right) = LLM_{GPT\text{-}4o}\left(Integrate\left(CA_{comp}, CoT_{reasoning}, SS_{search}\right)\right) \quad (2)$$

Where:

- $SA_{summary}$ (*Summarization Agent*): Synthesizes context-aware insights
- $CA_{comp}$ (*Computational Agent*): Processes microbial and clinical data to generate structured outputs
- $CoT_{reasoning}$ (*Chain-of-Thought reasoning*): Provides structured logical interpretive paths
- $SS_{search}$ (*Semantic Search*): Retrieves relevant contextual information using RAG from a vector database.
- $LLM_{GPT\text{-}4o}$ (*Large Language Model*): GPT-4o model that generates accurate, coherent summaries
- Integrate(·): Combines multiple sources into a unified semantic context

We propose a logic for CoT reasoning for the summarization agent in the following eight steps to ensure a comprehensive understanding of the input:

1. Patient Overview: Starts with basic patient demographics and general health status
2. Key Clinical Markers: Identifies important biomarkers, lab values, or health indicators
3. Gut Microbiome Profile: Analyzes the specific microorganisms present in the gut

4. Diversity Metrics Analysis: Evaluates biodiversity measures of the gut microbiome
5. Interactions and Mechanisms: Examines relationships between the microbiome and clinical markers
6. Descriptive Correlation: Identifies statistical relationships without implying causation
7. Machine Learning analysis and probabilistic assessment: ML models were used to evaluate the patterns and make probabilistic predictions
8. Final Comprehensive Descriptive Summary: Creates a holistic summary integrating all findings

*Classification Agent:* The classification agent integrates the output of the computational agent, the output of the summarization agent, CoT reasoning, our semantic search engine, and the GPT-4o-mini model for content classification. We used GPT-4o-mini for AD classification tasks because it efficiently processes extremely long texts, delivers high accuracy and reliability essential for medical diagnosis, and offers unmatched cost-effectiveness compared with similar models [30]. The formula for a classification agent is as follows:

$$CA_{class}\left(CA_{comp}, SA_{summary}, CoT_{reasoning}, SS_{search}, LLM_{GPT\text{-}} \right. \\ = LLM_{GPT\text{-}4o\text{-}mini}\left(Integrate\left(CA_{comp}, SA, CoT_{reasoning}, S\right.\right. \quad (3)$$

Where:

- $CA_{class}$ (*Classification Agent*): Responsible for generating Alzheimer's disease classification outputs
- $CA_{comp}$ (*Computational Agent*) contains structured data features (e.g., predictions and feature importance)
- $SA_{summary}$ (*Summarization Agent*): Provides high-level contextual insight
- $CoT_{reasoning}$ (*Chain-of-Thought reasoning*): Guides interpretability through logical step-by-step inference.
- $SS_{search}$ (*Semantic Search*): retrieves conceptually relevant knowledge from a vector-based index
- $LLM_{GPT\text{-}4o\text{-}mini}$ (*Large Language Model*): Optimizes for fast and efficient classification tasks
- Integrate(·) is a function that merges multiple contextual and data-driven sources into a unified representation

We propose the CoT reasoning logic for the classification agent in the following eight steps to ensure robustness in the classification task.

1. Historical Data Insights: Analyzes past patient data to establish baselines and trends
2. Diversity Metrics & Classification Refinement: Uses microbiome diversity measures to improve classification accuracy
3. Adaptive Threshold Decisioning: Determines dynamic thresholds for classification rather than fixed values
4. Handling Edge Cases & Misclassifications: Identifies and addresses outliers or difficult-to-classify cases
5. Comprehensive Summary of this Visit: Provides a complete analysis of the current patient data
6. SHAP Feature Importance: Uses SHAP values and ML outputs to understand the features that influence the model's predictions most.
7. Key Considerations for Prediction and Misclassification Adjustments: Identifying factors that might lead to misclassification and how to adjust for them.
8. Prediction Decision Rules: Establishes clear rules for making final classification decisions.

### B. Base LLM Models

This study utilized two OpenAI LLM models: GPT-4o for summarization and GPT-4o-mini for improved speed and cost efficiency. Their selection aligns with the design concepts of the summarization and classification agents. Using an eight-step CoT reasoning logic, the summarization agent is designed to handle more contextual data, including single-visit and longitudinal data. Each step corresponds to a relevant RAG output, explaining the meaning of each step in summarizing an individual's clinical conditions and microbiome profiles. The summarization task was designed to process approximately 100,000 tokens of text data at a time, given its superior ability to handle clinical text summarization [31]. Therefore, as a downstream counterpart to GPT-4o, GPT-4o-mini was proposed for focused binary classification tasks utilizing a dedicated CoT reasoning framework for classification. It processes data that has been preprocessed by computational and summarization agents powered by GPT-4o and is optimized for high-throughput classification over approximately 50,000 tokens per instance.

### C. Semantic Search Engine

The knowledge base comprised 76,751 full-text publications or abstracts programmatically retrieved in October 2024 from PubMed Central (PMC) using the NCBI Entrez system via E-utilities and Entrez Direct. Articles were identified based on user-defined query terms applied to specific fields, such as titles, abstracts, or text words, and retrieved in XML format using the efetch utility. The collected records were subsequently processed using an embedding model and indexed into two vector databases to support downstream retrieval tasks.

The publications were divided into 2,058,502 text chunks, each containing 2,000 characters, with a 20 percent overlap with the previous chunk to ensure continuity and preserve context. This approach minimizes the risk of losing critical information when analyzing the AD-related literature. The embedding model converts these chunks into high-dimensional vector representations, which are then stored in two separate vector databases optimized for semantic search and retrieval (FIGURE 5). These databases enable rapid querying of relevant AD research, supporting tasks such as disease classification, summarization, and hypothesis generation within the ADAM framework.

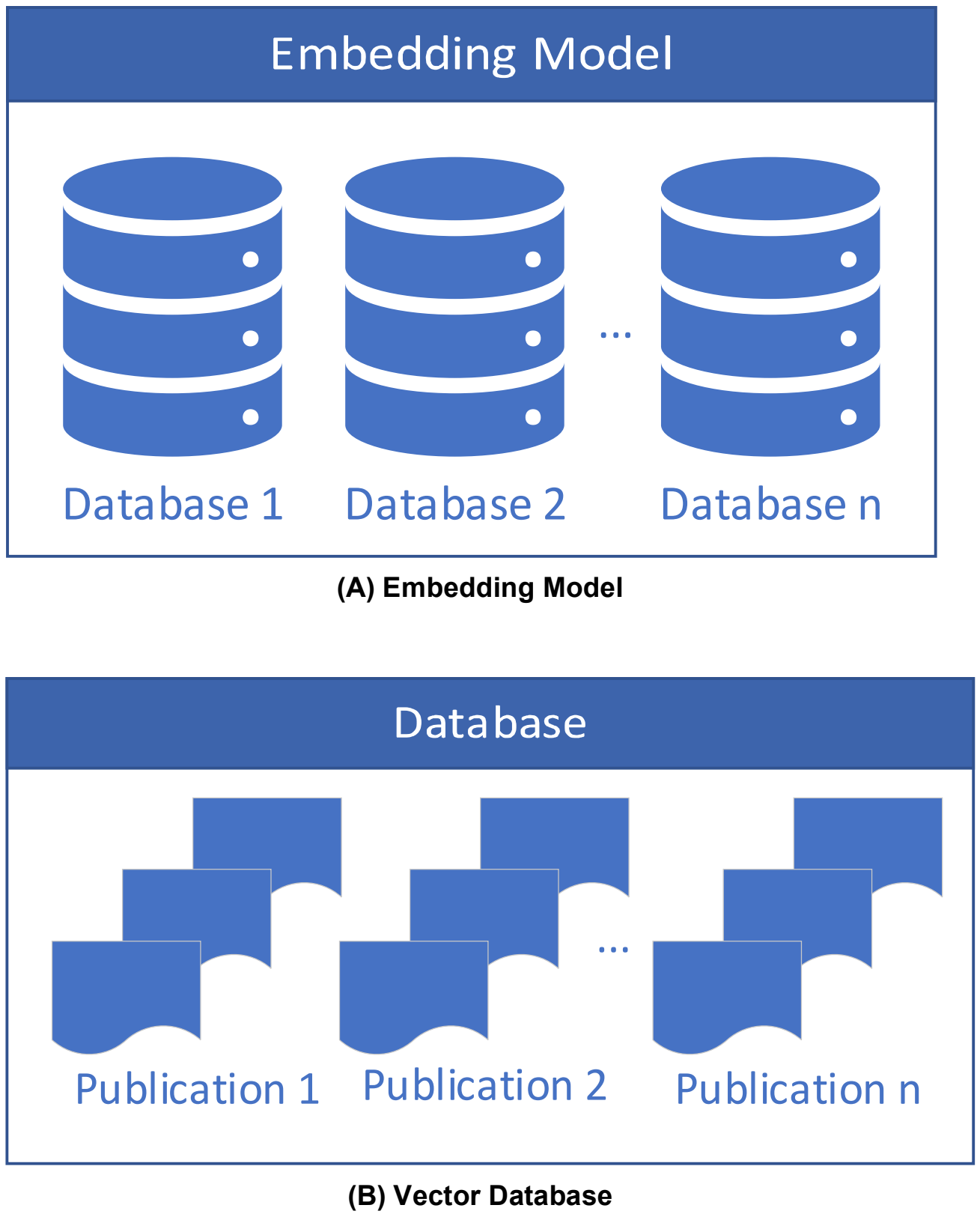

**FIGURE 5. Semantic Search Engine Architecture. Semantic Search Engine. (A) The embedding model converts documents into and out of numerical vectors. (B) Vector databases work in parallel with the semantic search engine that returns the most relevant information measured by cosine similarity. Inside a Vector Database. Each publication is split into 2000-character vector chunks, with a 20 percent overlap between the prior chunk of text and the next.**

*Publications*: This component serves as a source of existing literature and the reasoning logic of the ADAM, integrating 76,751 publications relevant to AD research at the time of this study. The keywords used for indexing included Alzheimer's, Bacterial Translocation, Gut-Brain Axis, Gut Microbiome, Immunosenescence, and Microbial Translocation (TABLE I).

TABLE I
PUBLICATIONS AND TEXT SEGMENT COUNTS BY TOPIC.

| Keywords | Publications | Segments |
|---|---|---|
| Alzheimer's disease | 62,478 | 1,591,441 |
| Gut Microbiome | 11,692 | 381,630 |
| Immunosenescence | 1,273 | 36,172 |
| Gut-Brain Axis | 1,308 | 49,259 |
| Total | 76,751 | 2,058,502 |

THE NUMBER OF PUBLICATIONS AND SEGMENTED TEXT UNITS ANALYZED IN THIS STUDY ACROSS KEY BIOMEDICAL TOPICS RELATED TO ALZHEIMER'S DISEASE AND MICROBIOME RESEARCH. THE DATASET INCLUDES OVER 80,000 PUBLICATIONS AND 2.1 MILLION SEGMENTS, MOST FOCUSED ON ALZHEIMER'S DISEASE AND THE GUT MICROBIOME.

Each publication $P_i$ is indexed by a set of relevant keywords $K_{i,j}$, forming a keyword-embedding vector $K(P_i)$:

$$\boldsymbol{K}(P_i) = \sum_{j=1}^{m} w_{i,j} \cdot \text{EmbeddingModel}(K_{i,j}) \quad (4)$$

Where:

$P_i$ represents the i-th publication.
$K_{i,j}$ is the j-th keyword associated with $P_i$.
$w_{i,j}$ is the weight assigned to $K_{i,j}$ based on its relevance to $P_i$.
$\boldsymbol{K}(P_i)$ is the aggregated embedding vector representing the keywords of $P_i$, facilitates its retrieval from the knowledge base.

The vector $\boldsymbol{K}(P_i)$ enables the efficient retrieval of publications based on semantic relevance to query keywords in AD research.

*Embedding Model*: The embedding model transforms textual data into embeddings, thereby allowing the framework to process and effectively retrieve relevant information. We chose text-embedding-ada-002 because of its high-cost effectiveness and proven success in text processing since its release in December 2022. To determine the most suitable embedding models for this study, we compared three available embedding models from OpenAI: text-embedding-ada-002 (hereafter referred to as ada-002), text-embedding-3-small (hereafter referred to as 3-small), and text-embedding-3-large (hereafter referred to as 3-large) to determine the most suitable embedding models for this study [32]. These three models were evaluated based on the following criteria: semantic richness, computational efficiency, storage requirements, cost-effectiveness, versatility, and adoption (TABLE II).

Furthermore, we applied these three embedding models to a small-scale test vector database derived from gut microbiome publications, where each text chunk contained 2,000 characters and had 20 percent overlap with the previous chunk, utilizing a cosine similarity threshold of 0.8. Our empirical findings indicated that the ada-002 model yielded more consistent results within the LLM and vector database configurations during the RAG retrieval process.

TABLE II.
EMBEDDING MODEL COMPARISON. SUMMARY OF KEY PERFORMANCE AND USABILITY ATTRIBUTES OF THE ADA-002 EMBEDDING MODEL, HIGHLIGHTING ITS STRONG SEMANTIC REPRESENTATION, HIGH COMPUTATIONAL EFFICIENCY, MODERATE STORAGE DEMANDS, AND BROAD ADOPTION ACROSS APPLICATIONS.

| Criterion | ada-002 (1,536 dims) | 3-small (1,536 dims) | 3-large (3,072 dims) |
|---|---|---|---|
| Semantic Richness | High | High | Very High |
| Computational Efficiency | High | Highest | Moderate to Low |
| Storage Requirements | Moderate | Moderate | High |
| Cost-effectiveness | High | Moderate | Lower |
| Versatility & Adoption | Proven, widely used | Newer, less proven | Newer, powerful, but resource-intensive |

Given an input text $\boldsymbol{T}$, the embedding model generates an embedding vector $\vec{\boldsymbol{E}}$ as follows:

$$\vec{\boldsymbol{E}} = \text{EmbeddingModel}(\boldsymbol{T}) \quad (5)$$

Where:

$\boldsymbol{T}$ represents the i-th publication.
$\vec{\boldsymbol{E}}$ is the resulting embedding vector in high-dimensional space.

This embedding vector $\vec{\boldsymbol{E}}$ captures semantic and contextual information from $\boldsymbol{T}$, facilitating effective retrieval and matching within the framework.

*Vector Databases*: When embedding publications into a vector database, the publication text is divided into 2000-character segments, with a 20 percent overlap between segments, to maintain the information flow and provide a more continuous and coherent representation of the text.

Given:
Segment length $\boldsymbol{s} = \boldsymbol{2000}$ characters
Overlap $\boldsymbol{o} = \boldsymbol{0.2} \times \boldsymbol{s} = \boldsymbol{400}$ characters
Effective step size = $\boldsymbol{s} - \boldsymbol{o} = \boldsymbol{1600}$ characters

$$\boldsymbol{p_i} = \boldsymbol{1} + (\boldsymbol{i} - \boldsymbol{1}) \times (\boldsymbol{s} - \boldsymbol{o}) \quad (6)$$

This formula calculates the starting position of the $\boldsymbol{i}$-th segment:
Segment 1 starts at position 1
Segment 2 starts at position 1601
Segment 3 starts at position 3201, and so on...

Total number of segments $\boldsymbol{n}$ required to cover the entire text $\boldsymbol{L}$, each new segment after the first adds $(\boldsymbol{s} - \boldsymbol{o}) = \boldsymbol{1600}$ new characters. Subtraction of $\boldsymbol{o}$ from the numerator accounts for its overlap with the last segment.

$$\boldsymbol{n} = \left\lceil \frac{\boldsymbol{L} - \boldsymbol{o}}{\boldsymbol{s} - \boldsymbol{o}} \right\rceil \quad (7)$$

For example:
If $\boldsymbol{L} = \boldsymbol{5800}$ characters
First segment covers positions 1-2000
Second segment covers positions 1601-3600
Third segment covers positions 3201-5200
Fourth segment covers the position of 5211-7200

Using the original formula: $\boldsymbol{n} = \lceil \frac{\boldsymbol{5800-400}}{\boldsymbol{2000-400}} \rceil = \lceil \frac{\boldsymbol{5400}}{\boldsymbol{1600}} \rceil = \lceil \boldsymbol{3.375} \rceil = \boldsymbol{4}$, as a result that 5800 characters need 4 segments.

This configuration incorporated a 20 percent overlap between consecutive segments to maintain continuity in the embeddings. Consequently, the vector databases in the semantic search engine comprised 2,136,895 discrete overlapping segments derived from 76,751 publications (TABLE I).

## IV. Hardware and Software

The experiments were conducted on an Ubuntu 24.04.2 LTS workstation equipped with an Intel® Core™ i9-10900X (20 threads), 128 GB RAM, and four NVIDIA GeForce RTX™ 3090 GPUs, providing a robust computing environment for LLM and machine learning tasks. The software stack was built on Python 3.10.14, utilizing XGBoost 2.1.3, and Scikit-Learn 1.5.2 for model training, with Optuna 4.1.0 handling hyperparameter optimization.

For LLM processing, the system integrates OpenAI 1.55.1, PandasAI 2.4.2, LangChain 0.3.8, and LangChain-Chroma 0.1.4. Additionally, Scikit-Bio 0.6.2, SciPy 1.10.1, NumPy 1.26.4, and Pandas 1.5.3 facilitated data processing and analysis. For visualization and interpretability, the setup included Matplotlib 3.7.5, Seaborn 0.12.2, and SHAP 0.46.0. This configuration ensures high computational efficiency and scalability for deep learning workflows.

## VI. Data Split and Seeding Policy

### A. Data Split Policy

We implemented two data-splitting policies: one for selecting the baseline model and the other for the ADAM framework. In the baseline model selection phase, the data were initially split at 75:25 by a unique Study ID and stratified according to the AD status. In the second phase, which involved the ADAM framework, 75% of the data was retained as the training or reference dataset. For testing, we randomly selected 15 positive and 15 negative cases from the 25% portion owing to hardware limitations. This approach allowed us to work efficiently with the LLM while maintaining statistical significance (n = 30).

### B. Seeding and Measures

We implemented two seeding strategies to support the different phases of the study: one for baseline model selection and the other for creating the ADAM framework. The initial phase involved applying 10 random seeds to identify the best-performing classifier based on the accuracy, AUC, and F1 score. The selected model was then integrated into ADAM, a language-model-based system designed to enhance the binary classification of AD. A broader evaluation was conducted using 30 random seeds to examine whether ADAM improved the mean F1 score and reduced the performance variance compared with the baseline. This seeding policy design ensured that the performance metrics in both phases were derived from stable and reproducible evaluations, allowing for a fair assessment of ADAM's added value over the baseline model.

## VII. Baseline Model Selection

We trained three classifiers–XGBoost, random forest, and logistic regression–on the AD data and identified the best-performing model as the baseline for this study. The model selection process involved two phases: feature selection and model training, both optimized using Optuna, which is a Bayesian-based hyperparameter-tuning framework [33]. To identify the most relevant features, we configured an XGBoost-based feature selector that was applied to all three machine learning models. The selected features are subsequently fed into the proposed models for further training and testing.

We implemented several performance metrics to evaluate the model's effectiveness, including accuracy, AUC, F1 score, and overall performance index. Each metric was calculated by averaging the results from 10 different random seeds, which helped capture the variability and enhance the robustness of the evaluation process. The results are shown in FIGURE 6 and TABLE III, which summarizes the comparative performance of each model across all metrics. By relying on multiple evaluation criteria, we aimed to assess not only how well each model classifies, but also how reliably and consistently it performs across different data splits, which is especially critical in clinical datasets where precision and balance are essential.

As shown in TABLE III, XGBoost consistently outperformed logistic regression and Random Forest across all key metrics. It achieved the highest average accuracy, AUC, and F1 score, demonstrating its superior discriminative ability and robustness. Furthermore, XGBoost exhibited the highest stability across different seeds, underscoring its reliability in managing complex and heterogeneous data related to AD. Owing to its strong performance and consistency, XGBoost was selected as the baseline model in this study. This served as the foundation against which the proposed ADAM framework was assessed, providing a high-performance and dependable benchmark for future model comparisons.

TABLE III.
BASE MODEL PERFORMANCE AVERAGED ACROSS 10 RANDOM SEEDS.

| Model | Accuracy | AUC | F1 | Mean |
|---|---|---|---|---|
| XGBoost | **0.769 ± 0.069** | **0.821 ± 0.061** | **0.651 ± 0.1** | **0.769 ± 0.069** |
| Random Forest | 0.742 ± 0.066 | 0.804 ± 0.087 | 0.603 ± 0.087 | 0.742 ± 0.066 |
| Logistic Regression | 0.735 ± 0.067 | 0.772 ± 0.1 | 0.626 ± 0.095 | 0.735 ± 0.067 |

THE CLASSIFICATION PERFORMANCE OF XGBOOST, RANDOM FOREST, AND LOGISTIC REGRESSION MODELS WAS EVALUATED USING MEAN ACCURACY, AUC, AND F1 SCORES WITH STANDARD DEVIATION. RESULTS REFLECT MODEL STABILITY AND PREDICTIVE POWER ACROSS REPEATED RUNS, WITH XGBOOST ACHIEVING THE HIGHEST AVERAGE PERFORMANCE ACROSS ALL METRICS.

## VIII. Results and Evaluation

### A. Evaluation Strategy

The evaluation strategy aimed to determine whether ADAM outperformed the baseline XGBoost model in binary classification tasks using the F1 score. The F1 score is directly affected by false negatives (FN) because it is the harmonic mean of precision and recall, with recall particularly sensitive to false negatives [34]. This is especially critical in medical applications, where false negatives can lead to severe consequences [35-37]. In the context of AD, false negatives can result in delayed diagnosis, leading to inadequate monitoring and treatment, loss of social and financial benefits, increased emotional distress for patients and caregivers, and ultimately, worse patient outcomes [38-40]. The mean F1 score quantifies the overall predictive performance, whereas the variance of the F1 score assesses the consistency and stability of the ADAM performance compared to XGBoost.

Each model was trained and evaluated over 30 independent runs with different random seeds, and the corresponding F1 scores were recorded. The statistical significance of the difference in mean F1 scores between ADAM and XGBoost was analyzed using the Mann-Whitney U test, a non-parametric test that does not require normality assumptions [41]. Additionally, an F-test was conducted to compare the variances in the F1 scores, providing insights into the performance stability of each model [42].

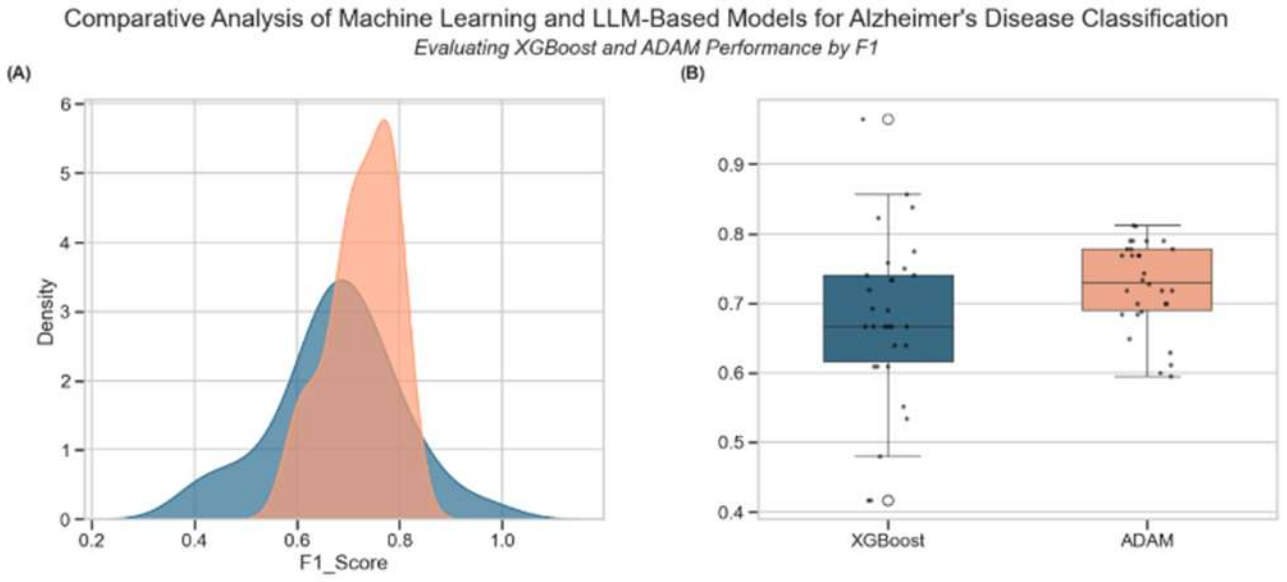

**FIGURE 6. Comparative Analysis of XGBoost and ADAM. (A) Density plots and (B) boxplots of F1 scores comparing XGBoost and the ADAM framework across multiple runs. ADAM demonstrates a higher median F1**

**score with reduced variance, suggesting more consistent and reliable classification performance for AD prediction.**

### *B. F1 Score and Variability Analysis*

As outlined above, the F1 score was selected for its suitability for imbalanced medical datasets and its sensitivity to false negatives, which are highly consequential in Alzheimer's disease classification.

Across 30 experimental runs, ADAM achieved a higher mean F1 score (0.7263) than XGBoost (0.6774), as determined by the Mann-Whitney U test ($p = 0.0418$), indicating a statistically significant improvement at the 95% confidence level. In addition, ADAM demonstrated a lower standard deviation (0.0632) than XGBoost (0.1217), as supported by Levene's test ($p = 0.0300$), indicating a more stable and consistent performance. The medium effect size (Cohen's $d = 0.5038$) further underscores the practical relevance of ADAM's superior and reliable classification capability.

The box plots and density distributions (FIGURE 6) demonstrate that ADAM maintains a narrower F1 score distribution (FIGURE 6(B)) with a higher median and fewer extreme values than XGBoost. Although both models exhibit similar central performance around 0.7, the density plot (FIGURE 6(A)) reveals that ADAM's F1 scores cluster more tightly around 0.75, whereas XGBoost displays a broader spread, ranging from 0.4 to nearly 1.0, indicating greater performance variability across different runs. This higher consistency in the ADAM performance suggests that it may offer more reliable predictions for AD classification (FIGURE 7).

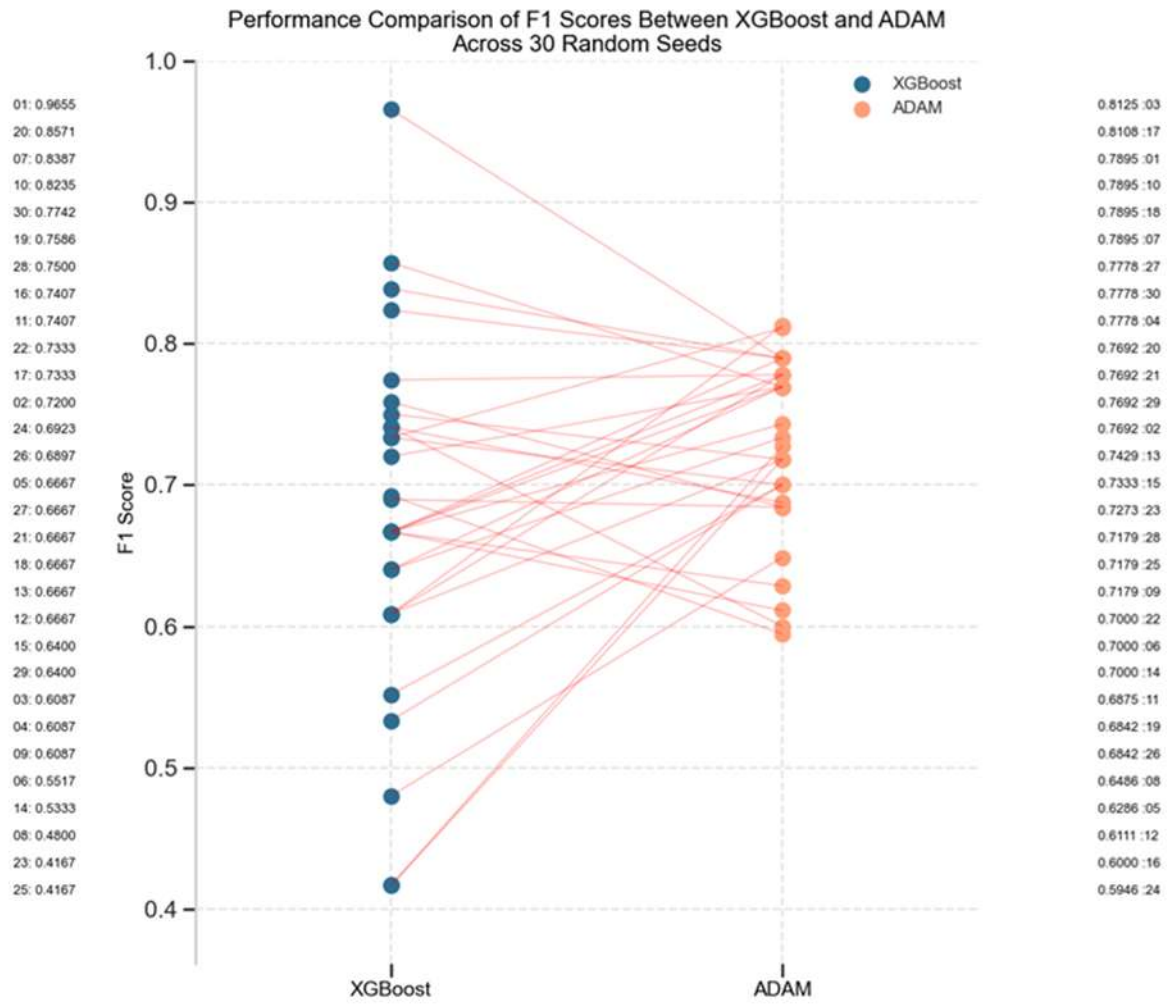

**FIGURE 7. Performance Comparison of F1 Scores Between XGBoost and ADAM. Line plot comparing F1 scores of XGBoost and ADAM across 30 random seeds. Each line connects paired runs for a given seed. ADAM consistently shows improved or more stable F1 performance relative to XGBoost, highlighting its robustness and reduced variability across repeated evaluations.**

These findings collectively demonstrate that ADAM significantly outperforms XGBoost, offering an improved average performance and greater consistency and stability across experimental runs.

## IX. Reporting Module

ADAM is not only a classifier but also a reporting system. Its classification is based on the combined results of the three agentic systems presented in textual form. Its logical reasoning design classifies the AD status while simultaneously generating analytical reports for each study subject. We listed two sample classification reports as follows:

### *A. Sample Reports*

TABLE IV presents two sample reports produced by ADAM, including one positive and one negative case. Owing to the length of the entire report, only the conclusion section is listed in each of the following reports. The full set of ADAM-generated reports, totaling over 900 case-specific classifications, is available at https://github.com/melhzy/ADAM/reporting. Two representative sample reports, positive and negative, were presented in a condensed format.

TABLE IV.
ADAM SAMPLE REPORTS: POSITIVE CLASSIFICATION AND NEGATIVE CLASSIFICATION. INTERPRETABILITY-FOCUSED SUMMARIES PRODUCED BY ADAM FOR ONE POSITIVELY AND ONE NEGATIVELY CLASSIFIED SAMPLE. EACH REPORT INCLUDES CLINICAL VARIABLES, MEDICATION USAGE, GUT MICROBIOME COMPOSITION, DIVERSITY METRICS, AND SHAP-BASED FEATURE ATTRIBUTIONS TO CLARIFY THE PREDICTED ALZHEIMER'S DISEASE STATUS.

| Sample of a Positive Classification - Conclusion **Prediction**: **Yes** - The prediction of Alzheimer's disease (AD) for Sample ID FB151 is strongly supported by multiple factors: | Sample of a Negative Classification - Conclusion **Prediction**: **No** - The prediction for Sample ID FB128 is 'No' for Alzheimer's disease (AD) based on a probability of 24.20%. This low probability is influenced by several key factors: |
|---|---|
| 1. **Clinical Indicators**: The patient is a 94-year-old female with a Clinical Frailty Scale score of 7, indicating severe frailty. This level of frailty is significantly associated with cognitive decline and AD progression. Additionally, the patient has a malnutrition score of 2, which places her at risk for nutritional deficiencies that can exacerbate cognitive decline. | 1. **Demographics and Clinical Background**: - The patient is 81 years old, male, with a moderate to severe frailty score of 6. This level of frailty is concerning but does not alone indicate a high probability of AD. – The malnutrition score of 2 indicates a risk of malnutrition, which can exacerbate cognitive decline but is not definitive for AD. |
| 2. **Polypharmacy**: The patient is on multiple medications (polypharmacy), which can alter gut microbiota composition and increase systemic inflammation, further contributing to cognitive health deterioration. | 2. **Clinical Features**: - The presence of seizure medications and oral corticosteroids suggests potential neurological comorbidities, which could elevate AD risk. However, the absence of cardiovascular risk factors (e.g., hypertension, high cholesterol) is protective against vascular contributions to cognitive decline. |
| 3. **Gut Microbiome Profile**: The microbiome analysis reveals an | 3. **Gut Microbiome Profile**: - The microbiome analysis shows |
| imbalance with elevated levels of pro-inflammatory species such as *Phocaeicola vulgatus* and *Clostridia bacterium*, while beneficial species like *Faecalibacterium prausnitzii* are absent. This dysbiosis is linked to increased inflammation and cognitive decline, which are critical factors in AD pathology. | a moderate abundance of protective species like *Faecalibacterium prausnitzii* (1.28183), which has anti-inflammatory properties. This is counterbalanced by the presence of *Neglecta timonensis* (0.74327), which is positively associated with AD risk. – The overall alpha diversity metrics (Shannon Index: 3.50, Simpson Index: 0.93) indicate a relatively balanced microbiome, which is generally protective against systemic inflammation and cognitive decline. |
| 4. **Diversity Metrics:** The Shannon Index of 2.98 indicates moderate diversity, but the presence of dominant pro-inflammatory species and high Bray-Curtis dissimilarity suggests significant deviations from a healthy microbiome. This imbalance can exacerbate systemic inflammation and cognitive decline. | 4. **SHAP Feature Importance**: - The top SHAP features include *Neglecta timonensis* (SHAP: +0.7978), Seizure Medications (SHAP: +0.7696), and *Faecalibacterium prausnitzii* (SHAP: -0.6193). The positive contributions from Neglecta and seizure medications suggest increased risk, while Faecalibacterium's negative contribution indicates a protective effect. |
| 5. **SHAP Feature Importance**: The SHAP analysis highlights that the malnutrition score, the presence of *Neglecta timonensis*, and the low levels of *Faecalibacterium prausnitzii* are the top contributors to the high probability of AD. These features align with known risk factors for AD. | 5. **Diversity Metrics**: - The Bray-Curtis dissimilarity indicates high dissimilarity from healthy controls, suggesting a distinct microbial composition that may reflect gut dysbiosis linked to AD. |

Analyses of samples FB151 and FB128 demonstrated that the gut microbiome composition significantly influenced AD classification. Sample FB151, from a 94-year-old female with severe frailty (CFS 7), exhibited clear microbiome dysbiosis, characterized by elevated pro-inflammatory species, absence of the beneficial *Faecalibacterium prausnitzii*, and lower diversity (Shannon index: 2.98), which led to a positive AD classification. In contrast, sample FB128, an 81-year-old male with moderate-to-severe frailty (score 6), maintained a more balanced microbiome with protective *F. prausnitzii* (1.28183) that counterbalanced the presence of AD-associated *Neglecta timonensis*, resulting in higher diversity metrics (Shannon, 3.50; Simpson, 0.93) and a negative classification despite a similar malnutrition risk (score 2). SHAP analyses for both patients highlighted the critical importance of *F. prausnitzii* as a protective factor and *Neglecta timonensis* as a risk-inducing factor, demonstrating how microbiome balance can mitigate clinical risk factors and potentially protect against AD development, even in advanced age and frailty.

## X. Discussion and Limitations

The ADAM framework demonstrates the potential for leveraging LLMs by integrating bioinformatics, machine learning, explainable AI, and relevant literature with biological laboratory data analytics. It is designed for typical biological experimental settings, where around 100 to 300 data points are generated per experiment. These datasets often contain noise that can impede the effectiveness of the biological analysis. When working with human-derived data, the levels of noise and variability are generally greater, presenting additional challenges for analysis and interpretation.

Despite these challenges, ADAM demonstrates that when combined with RAG and knowledge-grounded reasoning, large language models can effectively contextualize biological data, even in small samples or high-noise scenarios. This capability is particularly crucial in microbiome clinical research, where important patterns are often subtly dispersed across different data types. Employing explainable AI methods, such as SHAP values, enhances interpretability, allowing researchers to connect a model's predictions to specific features or insights drawn from the literature.

However, this study has some limitations that should be acknowledged. First, although the framework was only tested on small-scale clinical datasets, it was designed to support high-dimensional and large-scale biological data through agentic systems. Still, performance at this scale has yet to be fully validated, and future work is needed to assess scalability, particularly with complex multi-omics inputs. Second, although ADAM provided strong generalization capabilities in this study, further training, fine-tuning, and enhancements are strongly recommended to mitigate the impact of hallucinations when applied to other datasets or topics. Third, although the semantic search engine within ADAM leverages a large and diverse literature corpus to mitigate noise and variability in raw data and intermediate computational outputs, the effectiveness of this approach remains dependent on the coverage and quality of the embedded knowledge. Gaps or biases in literature may lead to overlooked findings or skewed interpretations, particularly in underrepresented research domains. Fourth, the two base LLMs are pretrained models from OpenAI, designed for more general purposes rather than detail-oriented biological studies, where subtle differences can have a significant impact. Finally, the reasoning logic in the summarization and classification agents is manually tuned, which may limit the adaptability across diverse datasets and require extensive domain expertise to adjust, optimize, and effectively process new data in various biological contexts.

In addition, the computational overhead associated with running multiple AI agents and real-time retrieval over large document bases may not be practical in all laboratory settings, particularly for those with limited infrastructure. Furthermore, although ADAM enables rapid data analysis and biological reasoning, expert interpretation is necessary to validate the findings and ensure scientific rigor.

## XI. Conclusion and Future Work

This study offers a thorough evaluation of the ADAM framework for classifying AD by comparing its performance with that of the widely used machine-learning baseline, XGBoost. By leveraging LLMs, explainable AI, and retrieval-augmented generation, ADAM achieved better predictive performance and showed greater consistency across repeated experimental runs.

Across 30 independent trials, ADAM consistently achieved a higher mean F1 score (0.7263) than XGBoost (0.6774), with a statistically significant difference confirmed by the Mann-Whitney U test ($p = 0.0418$). The lower standard deviation of the ADAM F1 scores (0.0632 vs. 0.1217) and its favorable distribution characteristics further underscore its stability and robustness. The F1 variance of XGBoost (0.0148) indicated that XGBoost shows 3.71 times more variability than ADAM (0.0040), as supported by Levene's test ($p = 0.0300$). These performance characteristics are particularly important in the medical domain, where model stability and reproducibility are essential for building trust in diagnostic tools and supporting consistent decision-making in patient care.

The ADAM framework offers a promising direction for integrating AI-driven inference with biological and clinical data, providing higher accuracy, greater interpretability, and operational stability. These results support its potential for adoption in biomedical research environments and clinical decision support systems, particularly in settings characterized by small, noisy, or imbalanced datasets.

Overall, ADAM addresses the crucial gap between experimental biology and agentic AI. Future enhancements, such as domain-specific LLMs and fine-tuning, expanded knowledge integration, adaptive reasoning logic using reinforcement learning, and real-time validation mechanisms, are essential for broader adoption and translational impact. These developments may ultimately enable a foundational LLM trained and explicitly tuned for AD research, driving this line of research toward the realization of physical AI systems capable of interacting with and reasoning about complex biological data in real-time.

## Code Availability

The complete code that supports all the findings and analyses presented in this study is available at https://github.com/melhzy/ADAM. This repository includes data preprocessing scripts, model training scripts, evaluation scripts, and documentation detailing the setup procedures and dependencies. The code is distributed under the MIT License, permitting reuse and modification with appropriate attribution. Please use the repository issue tracker or contact the corresponding author, Dr. Ziyuan Huang, for any questions or concerns. For inquiries related to Alzheimer's and its microbiome, please contact Dr. John P. Haran. For computational biology inquiries, please contact Dr. Bucci Vanni.

## Data Availability

All data are available under BioProject accession number PRJNA529586 at NCBI and are further described in the Supplementary Data at doi.org/10.1128/mBio.00632-19 at mBio.

## Ethics Approval

This prospective cohort study was approved by the Institutional Review Board (IRB) of the University of Massachusetts Medical School (Docket H00010892).

## ACKNOWLEDGMENT

The authors would like to express their sincere gratitude to the members of the Bucci Lab at UMass Chan Medical School for their valuable contributions to this work. Their insightful feedback, technical support, and collaborative spirit significantly enhanced the development and refinement of this manuscript. We are especially thankful for their expertise and dedication throughout the research process.

The ADAM framework automatically generated the sample reports provided in TABLE IV, which leverages GPT-4o and GPT-4o-mini as its backend LLMs to process and analyze clinical and microbiome data for Alzheimer's disease summarization and classification.